\pdfoutput=1

\documentclass[11pt]{article}

\usepackage{EMNLP2023}

\usepackage{times}
\usepackage{latexsym}

\usepackage[T1]{fontenc}

\usepackage[utf8]{inputenc}

\usepackage{microtype}

\usepackage{inconsolata}
\usepackage{graphicx}
\usepackage{subfigure}
\usepackage{listings}
\usepackage{booktabs}
\usepackage{multirow}
\usepackage{tabularx}
\usepackage{xtab}

\title{\raisebox{-0.27\height}{\includegraphics[height=1.5\baselineskip]{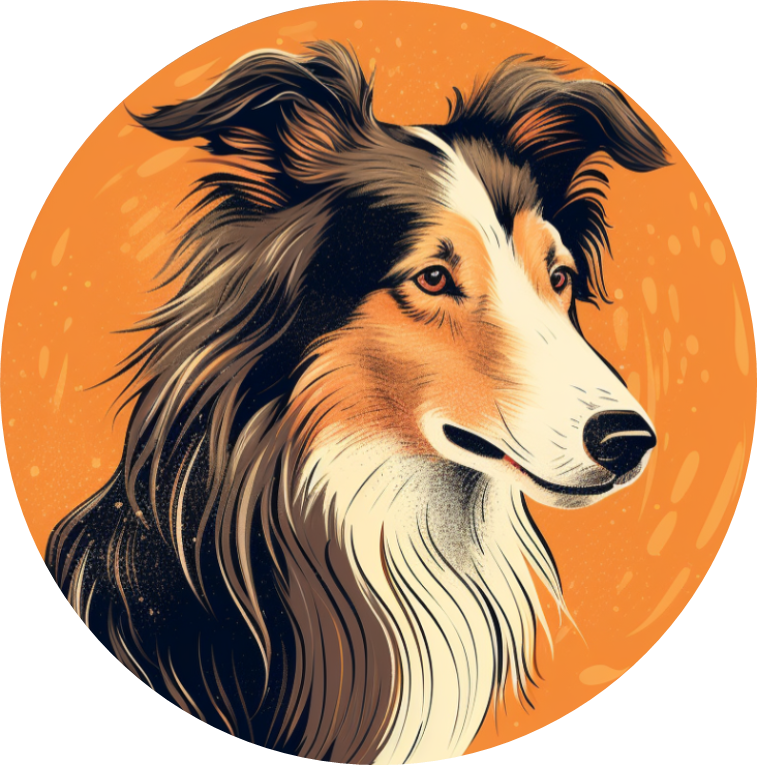}} \  CoLLiE: Collaborative Training of Large Language Models\\ in an Efficient Way}

\author{
Kai Lv\textsuperscript{1}\thanks{\ \ Equal contribution.
}\,\:\thanks{\ \ Work done during internship at Shanghai AI Lab.} , Shuo Zhang\textsuperscript{1}$^{*}$\footnotemark[2]\;, Tianle Gu\textsuperscript{2}, Shuhao Xing\textsuperscript{1}, Jiawei Hong\textsuperscript{1}, Keyu Chen\textsuperscript{1} \\
\textbf{Xiaoran Liu\textsuperscript{1}, Yuqing Yang\textsuperscript{1}, Honglin Guo\textsuperscript{1}, Tengxiao Liu\textsuperscript{1}, Yu Sun\textsuperscript{1}}\\
\textbf{Qipeng Guo\textsuperscript{1}, Hang Yan\textsuperscript{2}\thanks{\ \ Corresponding authors.}\;, Xipeng Qiu\textsuperscript{1}\footnotemark[3]}\\
\textsuperscript{1} School of Computer Science, Fudan University,
\textsuperscript{2} Shanghai AI Laboratory \\
\textsuperscript{1}\{klv21, szhang22, hongjw21, kychen22, liuxr22\}@m.fudan.edu.cn \\
\textsuperscript{1}\{shxing, hlguo20, qpguo16, xpqiu\}@fudan.edu.cn, 
\textsuperscript{2}\{gutianle, yanhang\}@pjlab.org.cn
}

\begin{document}
\maketitle
\begin{abstract}
Large language models (LLMs) are increasingly pivotal in a wide range of natural language processing tasks. Access to pre-trained models, courtesy of the open-source community, has made it possible to adapt these models to specific applications for enhanced performance. However, the substantial resources required for training these models necessitate efficient solutions. 
This paper introduces CoLLiE, an efficient library that facilitates collaborative training of large language models using 3D parallelism, parameter-efficient fine-tuning (PEFT) methods, and optimizers such as Lion, Adan, Sophia, LOMO and AdaLomo. 
With its modular design and comprehensive functionality, CoLLiE offers a balanced blend of efficiency, ease of use, and customization. CoLLiE has proven superior training efficiency in comparison with prevalent solutions in pre-training and fine-tuning scenarios. 
Furthermore, we provide an empirical evaluation of the correlation between model size and GPU memory consumption under different optimization methods, as well as an analysis of the throughput. Lastly, we carry out a comprehensive comparison of various optimizers and PEFT methods within the instruction-tuning context. CoLLiE is available at \href{https://github.com/OpenLMLab/collie}{https://github.com/OpenLMLab/collie}.
\end{abstract}

\section{Introduction}
\begin{figure}[t]
    \centering
    \includegraphics[width=0.48\textwidth]{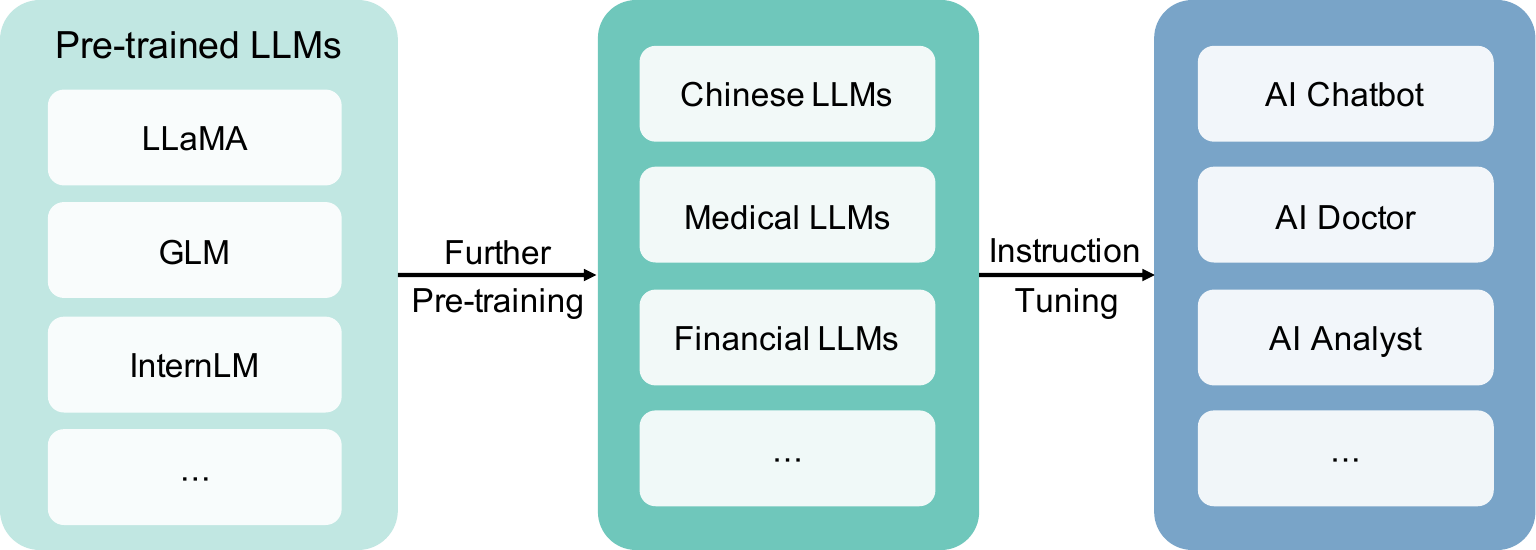}
    \caption{The two stages of training pre-trained language models, during which CoLLiE exhibits efficiency.}
    \vspace{-10pt}
    \label{fig:llm_training_stage}
\end{figure}

Large language models (LLMs) have demonstrated remarkable abilities across various natural language processing tasks and showcased potential as intelligent assistants. Thanks to the vibrant open-source community, multiple excellent large language models' weights are accessible, including  OPT~\citep{opt}, BLOOM~\citep{bloom}, LLaMA~\citep{llama}, etc. 
Despite the impressive general capabilities of pre-trained LLMs, training for particular application scenarios can lead to even more outstanding performance. As shown in Figure~\ref{fig:llm_training_stage}, the training process can be divided into two stages: 1. \textbf{Further pre-training}, which supplements specific domain knowledge and expands the vocabulary to enhance tokenization efficiency; 2. \textbf{Instruction-tuning}, which adapts the model to downstream tasks and improves its instruction-following ability.

With the scaling of language models, the resources required for training have increased substantially, making it infeasible to train the entire model on a single GPU. Model parallelism addresses this issue by partitioning the model across different GPUs, distributing the training workload among these GPUs. This can be achieved through three methods: tensor parallelism (TP, \citet{megatron}), pipeline parallelism (PP, \citet{gpipe,pipedream}), and stage 3 of zero redundancy optimizer (ZeRO-3, \citet{zero}). In addition, during the instruction-tuning stage, there are resource-efficiency and training-effectiveness trade-off approaches~\cite{compare_peft_full-tuning}: parameter-efficient fine-tuning (PEFT) methods~\cite{peft_overview}. These methods selectively choose or add a few parameters for training, effectively reducing the GPU memory required to train large language models.

In this context, we introduce CoLLiE, an easy-to-use library for \textbf{Co}llaborative training of \textbf{L}arge \textbf{L}anguage models \textbf{i}n an \textbf{E}fficient way. 
The library not only integrates the previously mentioned three parallelism strategies and PEFT methods, but also implements efficient optimizers such as Lion~\cite{lion}, Adan~\cite{adan}, Sophia~\cite{sophia}, LOMO~\cite{lomo} and AdaLomo~\cite{adalomo}.
We have restructured multiple mainstream open-source models to support TP and PP and incorporated FlashAttention~\cite{flashattn,flashattn2} to further boost efficiency, while retaining interfaces similar to HuggingFace models within \verb|CollieModel| class.
Training efficiency is one of the most distinctive feature of CoLLiE, boasting a significantly higher training throughput compared to current popular solutions.
CoLLiE also offers a wide range of functionalities, including data preprocessing, model training, checkpoint saving and monitoring and evaluation during training process. 
CoLLiE's modular design allows for flexible combinations of parallelism strategies, PEFT methods, and training hyperparameters, which can be configured simply by modifying the \texttt{CollieConfig} class.
Furthermore, CoLLiE is purposefully designed with extensibility, providing customizable functionalities. 
In summary, CoLLiE offers a comprehensive solution that caters to the needs of both beginners and experienced professionals. 
Our contributions can be summarized as follows:
\begin{itemize}
    \item We introduce CoLLiE, an efficient and easy-to-use library for collaborative training of large language models.
    \item We empirically provide the relationship
    between model size and the actual GPU memory consumption using different optimization methods in real training scenarios.
    \item We compared the throughput of CoLLiE and the current prevailing solutions in (further) pre-training and fine-tuning scenarios, and CoLLiE demonstrates higher efficiency.
    \item We conducted a comprehensive comparison of different optimizers and PEFT methods in the context of instruction-tuning.
\end{itemize}

\section{Background}
\paragraph{PEFT Methods}
There has been a rise in using parameter-efficient fine-tuning (PEFT) techniques to adapt models for instruction-tuning by adjusting partial parameters.
One of the early success is adapter tuning~\cite{adapter-tuning}, which inserts trainable neural modules into transformers layers while keeping the original model unchanged. In line with adapter tuning, LoRA~\cite{lora} reparameterizes the dense layers and only updates low rank matrices while introducing no latency during inference.
Prefix-tuning~\cite{prefix-tuning} trains a task specific prefix prepended to each layer of the transformer encoder and achieves comparable performance with full parameter fine-tuning on generative tasks. Similarly, prompt-tuning~\cite{prompt-tuning} simplifies the additional prefix to the input embeddings, and only updates the parameters corresponding to the prompts. 

While the PEFT library~\cite{peft} implements these algorithms at the model level, it relies on HuggingFace models and lacks a comprehensive functionality, particularly the necessary integration with model parallelism to facilitate training of extremely large models.

\paragraph{Parallelism Strategies}
Parallelism strategies refer to the methodology of utilizing multiple GPUs to execute training or inference tasks. 
Data parallelism involves distributing the input data to different GPUs for computation. However, each GPU stores an identical copy of the optimizer state and model weights, which limits the maximum model size that can be trained with data parallelism to that of a single GPU. 
To mitigate this redundancy, \citet{zero} proposes a parallelism strategy in the three stages of ZeRO, evenly partitioning the optimizer states, gradients, and weights across different GPUs.
Tensor parallelism also partitions the weights evenly, while it varies the approach to partition and communicate. Specifically, whereas ZeRO-3 gathers the weight matrices, tensor parallelism all reduces the intermediate computational results. 
Pipeline parallelism partitions the model by layers across GPUs, requiring communication only between the layers at the split points. This strategy yields the least communication overhead.

Existing toolkits, such as HuggingFace's Trainer~\cite{transformers} and LMFlow~\cite{lmflow}, choose ZeRO-3 as parallel method. ZeRO-3 is preferred because it does not impose specific requirements on the model structure, allowing direct usage of HuggingFace models. However, it exhibits lower throughput compared to the combination of TP and PP in scenarios involving large batch size pre-training or constrained communication.
CoLLiE supports the hybrid application of data parallelism, tensor parallelism, and pipeline parallelism, collectively termed as 3D parallelism, with the parallel sizes adjustable via \texttt{CollieConfig}.

\section{CoLLiE}

\begin{figure*}[t]
    \centering
    \subfigure[Architecture of CoLLiE. The blocks represent different modularly designed classes or the outputs of the \texttt{Trainer}.]{
        \includegraphics[height=5.4cm]{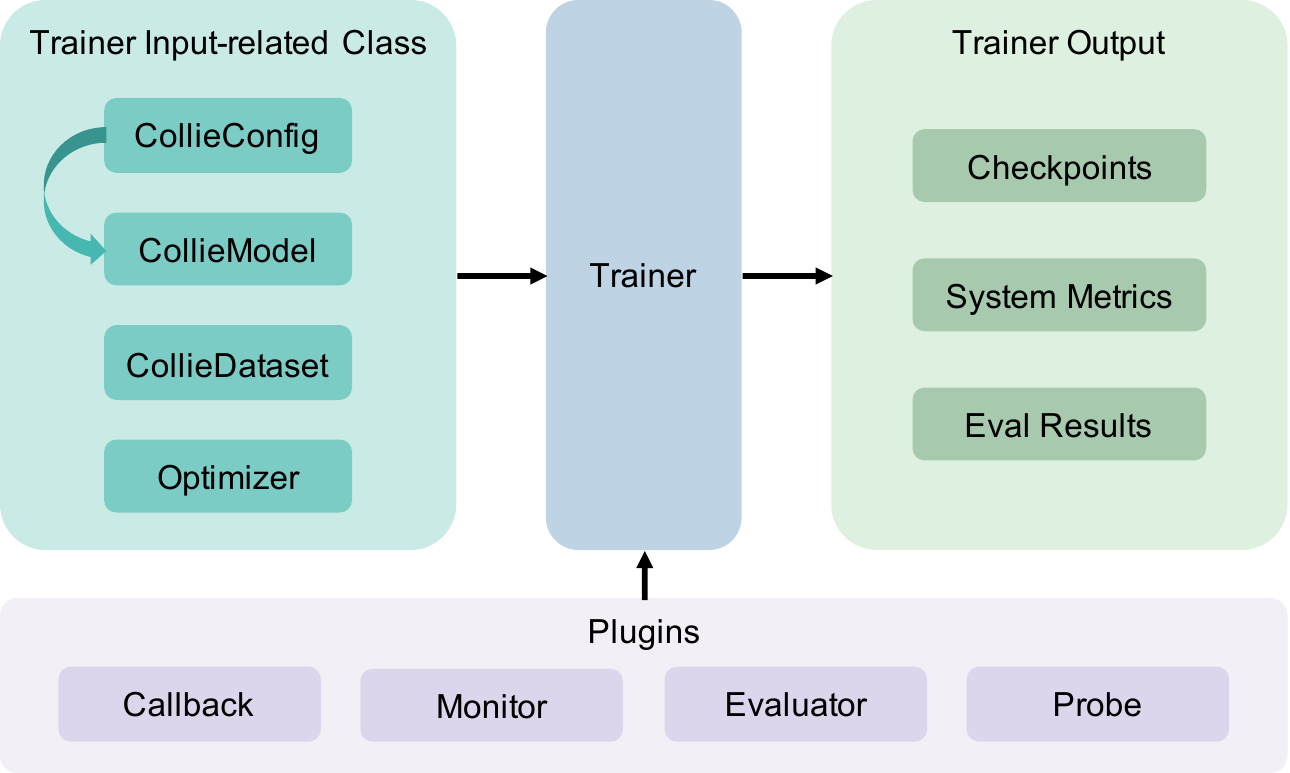}
        \label{fig:overview}
    }
    \hspace{0.025cm}
    \subfigure[Features of CoLLiE. CoLLiE supports a collaborative suite of high-efficiency optimization features.]{
        \includegraphics[height=5.4cm]{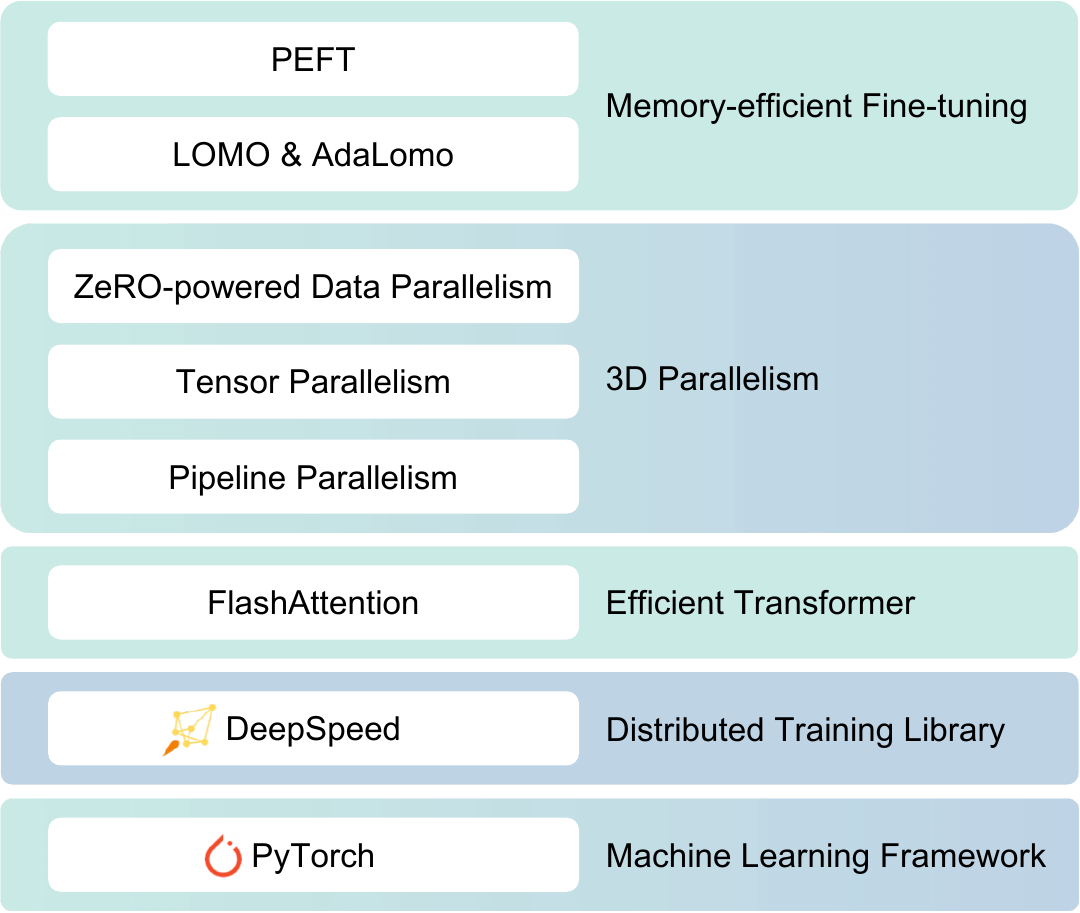}
        \label{fig:features}
    }
    \caption{Overall architecture and features of CoLLiE. Features in (b) are color-coded to match corresponding part in (a), indicating where each feature is implemented.}
\end{figure*}

In this section, we will introduce the implementation and features of CoLLiE. Figure~\ref{fig:overview} presents an overview of the Collie's overall structure, centered around the \texttt{Trainer} class. 
The \texttt{CollieConfig}, \texttt{CollieModel}, \texttt{CollieDataset}, and \texttt{Optimizer} classes serve as inputs to the \texttt{Trainer}. 
CoLLiE also provides a set of convenient plugins for the \texttt{Trainer}, including \texttt{Callback}, \texttt{Monitor}, \texttt{Evaluator}, and \texttt{Probe}, enabling users to customize the training process. Depending upon the configurations and selected plugins, the \texttt{Trainer} performs the training process, saves model checkpoints, and records system metrics, including loss, throughput, and evaluation results.

As shown in Figure~\ref{fig:features}, based on PyTorch~\cite{pytorch} and DeepSpeed~\cite{deepspeed}, CoLLiE employs a collaborative approach using various techniques to facilitate the more efficient training of large language models. Specifically, CoLLiE integrates FlashAttention to enhance efficiency. It implements ZeRO-powered DP, TP, and PP to support 3D parallelism. Additionally, LOMO, AdaLomo and PEFT methods are incorporated to realize memory-efficient fine-tuning approaches.

Appendix~\ref{sec:example} provides a brief tour demonstrating how to use CoLLiE for training.

\definecolor{deepgreen}{rgb}{0,0.5,0}

\lstdefinestyle{mystyle}{
    language=Python,
    basicstyle=\ttfamily\small,
    keywordstyle=\color{blue},
    commentstyle=\color{gray},
    stringstyle=\color{deepgreen},
    numbers=left,
    numberstyle=\tiny\color{gray},
    stepnumber=0,
    numbersep=5pt,
    frame=single,
    breaklines=true,
    breakatwhitespace=true,
    tabsize=4
}

\subsection{Collaborative Tuning Methods}

\subsubsection{3D Parallelism}
While distributed training frameworks such as DeepSpeed and Colossal-AI~\cite{colossalai} support 3D parallelism, models in HuggingFace can only opt for ZeRO-3 for model parallelism due to structural constraints. To fully support 3D parallelism and meet the distributed training needs under different scenarios, CoLLiE rewrites the models using Megatron-LM~\cite{megatron} and restructures them according to DeepSpeed's structure requirements for pipeline models. In the rewriting process, we have maintained the interface to be essentially consistent with the HuggingFace models, and have allowed the direct use of the \texttt{from\_pretrained} method to load pre-trained models from the HuggingFace hub. This approach significantly reduces the learning curve for users.

\subsubsection{Parameter-efficient Fine-tuning}
The PEFT library implements state-of-the-art PEFT methods at the model level, but lacks distributed training capabilities. CoLLiE has integrated the PEFT library into \texttt{CollieModel}, and made necessary patches to enable distributed training.

\subsubsection{Efficient Optimizers}
In addition to the popular AdamW~\cite{adam} optimizer, several other optimizers have been proposed for the purpose of saving memory, improving optimization results, or accelerating convergence. The implementation of the optimizers in CoLLiE is decoupled from other parts, and incorporates a variety of novel optimizers including Adan, Lion, Sophia, LOMO and AdaLomo. The effectiveness of these optimizers in training large language models is verified and compared in Section~\ref{sec:assessment}.

\subsection{Models}
In addition to the above-mentioned model implementations, CoLLiE has also replaced the naïve self-attention implementation with FlashAttention. Given that FlashAttention has strict requirements regarding hardware and CUDA versions, for users without newer training equipment, we have added the `use\_flash' option to the \texttt{CollieConfig} to allow for one-click disabling of FlashAttention usage. Currently, CoLLiE has implemented a variety of language models, including but not limited to LLaMA, InternLM~\cite{internlm}, ChatGLM~\cite{glm}, and MOSS~\cite{moss}, with the intention to support more models in the future.

\subsection{Configuration}
CoLLiE offers a unified class, \texttt{CollieConfig}, to manage configurations including model config, parallelism strategy, DeepSpeed configuration, PEFT configuration, and training hyperparameters. 
Based on the contents of \texttt{CollieConfig}, \texttt{CollieModel} will automatically adjust the partitioning of model parameters and the structure of the model, and the \texttt{Trainer} will modify the training process. 
Through \texttt{CollieConfig}, users can conveniently combine different pre-trained language models, fine-tuning methods, and hyperparameters.

Model config refers to parameters that describe the model structure, such as \texttt{hidden\_size}, \texttt{num\_attention\_heads}, and \texttt{num\_hidden\_layers}. The model config is fixed for pre-trained language models, and we provide a \texttt{from\_pretrained} interface, identical to HuggingFace's, to initialize model config. Users can also specify the model config to customize their models, intended for training from scratch without the use of pre-trained models. Below is a code example of downloading the model config from the HuggingFace hub, initializing the \texttt{CollieConfig}, and setting up to use FlashAttention.
\begin{lstlisting}[style=mystyle]
config = CollieConfig.from_pretrained(
    'meta-llama/Llama-2-7b-hf'
)
config.use_flash = True
\end{lstlisting}

\texttt{CollieConfig} streamlines the setup for 3D parallelism as followings. CoLLiE will automatically configure the distributed environment and partition the parameters according, relieving users from managing the complexities of distributed training. The number of GPUs required for training is equal to the product of the three parallelism sizes.
\begin{lstlisting}[style=mystyle]
config.dp_size = 1
config.tp_size = 8
config.pp_size = 2
\end{lstlisting}

CoLLiE implements distributed training based on DeepSpeed, and DeepSpeed-related configurations can be set via \texttt{ds\_config}. The configurations related to PEFT methods can also be set via \texttt{peft\_config}. Below is an example for mixed-precision training with FP16 and LoRA.
\begin{lstlisting}[style=mystyle]
config.ds_config = {
    'fp16': {'enabled': True}
}
config.peft_config = LoraConfig(
    r=4,
    lora_alpha=32,
    target_modules=['q_proj', 'v_proj'],
    bias='none',
    task_type='CAUSAL_LM'
)
\end{lstlisting}

The training hyperparameters can also be configured through \texttt{CollieConfig}.
Loading \texttt{CollieConfig} from a file is supported and we provide a convenient Command Line Interface (CLI) to generate the required configuration file.

\subsection{Dataset}
To facilitate data processing, CoLLiE provides three \texttt{Dataset} classes for training, evaluation of generation tasks, and evaluation of classification tasks respectively:  
\texttt{CollieDatasetForTraining},  \texttt{CollieDatasetForGeneration},  \texttt{CollieDataset
ForClassification}. These three classes can either read data from a JSON file or a list of dictionaries, process it, and store the results on disk for direct reading next time.

\texttt{CollieDatasetForTraining} accepts two forms of input, one with the field ``\texttt{text}'', and the other with fields ``\texttt{input}'' and ``\texttt{output}''. The loss of tokens in the field ``\texttt{text}'' or ``\texttt{output}'' will be computed, corresponding to pre-training and instruction-tuning tasks, respectively.
\texttt{CollieDatasetForGeneration} and \texttt{CollieDatasetForClassification} both inherit from the \texttt{CollieDatasetForTraining} class, serving as the datasets for generation tasks and classification tasks, respectively.  The \texttt{CollieDatasetForGeneration} can accept ``\texttt{text}'' as a required field and ``\texttt{target}'' as an optional field. The model generates output based on the ``\texttt{text}'' and the ``\texttt{target}'' is used to compute metrics in \texttt{Evaluator}.
On the other hand, \texttt{CollieDatasetForClassification} can accept ``\texttt{input}'', ``\texttt{output}'', and ``\texttt{target}'' fields. The ``\texttt{input}' represents the question, ``\texttt{output}'' includes all possible options, and ``\texttt{target}'' indicates which option should be chosen.




\subsection{Controller}
In this section, we will introduce three modularly designed classes centered around \texttt{Trainer}. \texttt{Trainer} calls the \texttt{Evaluator} and \texttt{Server} classes unidirectionally to serve the purposes of evaluation or manual probing of the model during training.

\subsubsection{Trainer}
Distributed training, including the initialization of the distributed environment, training loop, and the saving of model weights and checkpointing, can be complex. CoLLiE provides a \texttt{Trainer} to alleviate this burden on users. The \texttt{Trainer} wraps the relatively fixed training loop and offers multiple interfaces for users to further customize the training process. These include the \texttt{train\_fn} function that obtains output based on a given batch of input and the \texttt{loss\_fn} function that obtains loss based on the batch and output from \texttt{train\_fn}. Moreover, CoLLiE offers several plugins to enrich functionality. 

\noindent\textbf{Monitor}
The \texttt{Monitor} class tracks various metrics such as loss, learning rate, throughput, and memory usage during the training process, and records them to Tensorboard, WandB, or local CSV files.

\noindent\textbf{Callback}
The \texttt{Callback} class can be invoked at various callback points during the training process, allowing users to customize the training loop. CoLLiE has implemented callbacks that save model weights and training checkpoints when necessary, or load the model weights of the best evaluated results after training. These callbacks are all implemented based on the same base class. Users can inherit from this base class and override different methods to choose the callback timing and actions.

\subsubsection{Evaluator}
The \texttt{Evaluator} class is used in conjunction with the \texttt{Metric} class for assessing model performance. We implemented three types of \texttt{Evaluator}, intended for generation tasks, classification tasks, and perplexity assessment, by subclassing the \texttt{Evaluator} base class and overriding the \texttt{eval\_fn} method. The return value of the \texttt{eval\_fn} method in the \texttt{Evaluator} is accepted as input by the \texttt{update} function of the \texttt{Metric} class. The \texttt{Metric} class's \texttt{update} method updates the variables necessary for calculating the metric after processing each batch from the evaluation dataset. After the evaluation dataset is fully processed, the \texttt{get\_metric} function is employed to compute the metric. The \texttt{Evaluator} class can either be provided to the \texttt{Trainer} for assessment during the training process or it can evaluate the model independently without the dependency on the \texttt{Trainer}.

\subsubsection{Server}
The \texttt{Server} class offers web-based, interactive and streaming generated sequences feature, enabling users to conveniently deploy trained models for web-based use, as well as manually probe model performance during training. The \texttt{DataProvider} class supplies asynchronous inference data for \texttt{Server} as a subprocess. When the \texttt{Server} is integrated into the \texttt{Trainer}, users can input prompts via the web interface. Once the current batch training is completed, an output will be generated based on the user's input prompt and returned to the web interface for user's review.

\subsection{Documentation}
We provide API documentation and easily understandable tutorials\footnote{\href{https://openlmlab-collie.readthedocs.io/zh\_CN/latest}{https://openlmlab-collie.readthedocs.io/zh\_CN/latest}} for users who are new to CoLLiE and distributed training. Comprehensive code examples\footnote{\href{https://github.com/OpenLMLab/collie/tree/main/examples}{https://github.com/OpenLMLab/collie/tree/main/examples}} including vocabulary expansion, instruction-tuning, and downstream tasks such as summary and translation are also available.

\section{Evaluation}

\definecolor{mygray}{gray}{0.6}

\label{sec:eval}
\subsection{Memory Requirement}
\begin{figure}
    \centering
    \includegraphics[width=0.45\textwidth]{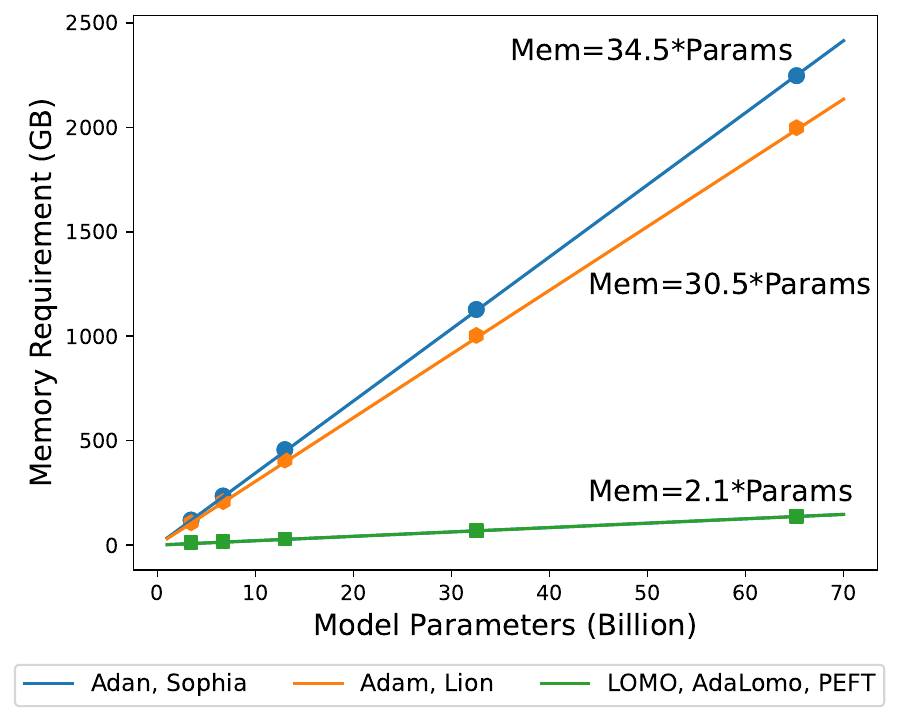}
    \caption{Memory requirements when training models with different parameters under various configurations.}
    \label{fig:mem_req}
\end{figure}
While \citet{zero} estimates the total GPU memory required for model training as 18 times the number of model parameters in bytes, more GPU memory is consumed in reality. This is because this estimation only considers memory used by parameters, gradients, and the optimizer states, and neglects other components such as activation values and buffers used for communication. 

In this section, we profile the actual memory requirements for training models under different configurations to facilitate users in more accurately estimating the model size that their devices can train.
As depicted in Figure~\ref{fig:mem_req}, the most commonly used Adam optimizer requires 30.5 times the amount of memory relative to the model parameters, which is consistent with Lion. Adan and Sophia optimizers use 4 times more memory for intermediate variables when updating parameters, amounting to 34.5 times the parameter size. LOMO and AdaLomo, without storing any optimizer state or gradient, only require 2.1 times the parameter size in memory, almost all of which is consumed by the half-precision parameters. PEFT methods, which update only a small proportion of parameters, have a memory usage similar to LOMO or AdaLomo.

\subsection{Throughput Analyses}
\begin{figure}
    \centering
    \includegraphics[width=0.48\textwidth]{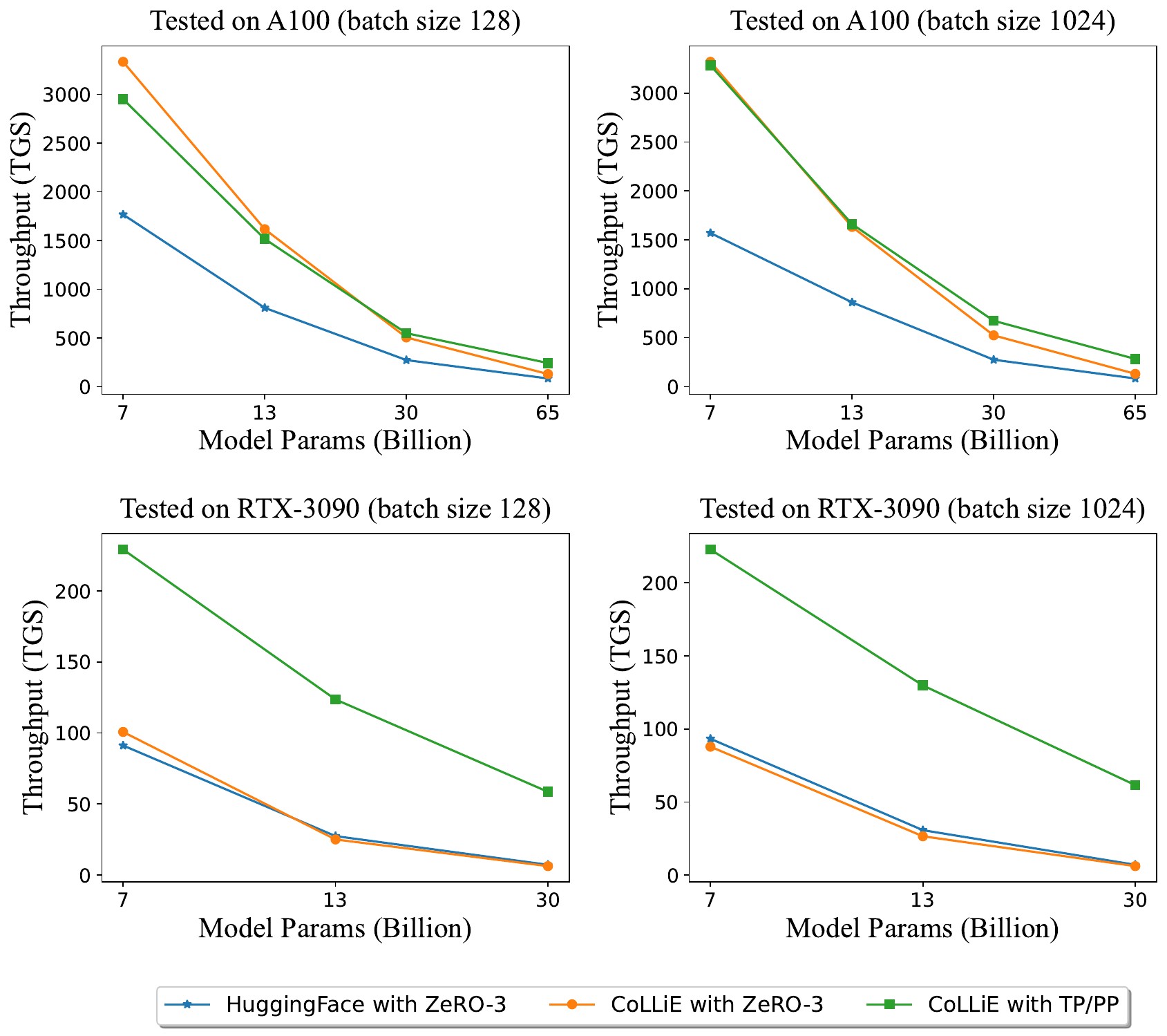}
    \caption{Throughput tested on A100 and RTX-3090.}
    \label{fig:throughput}
\end{figure}

We take HuggingFace models with ZeRO-3 as a baseline to analyse the throughput of CoLLiE during pre-training (with batch size of 1024) and fine-tuning (with batch size of 128). The corpus we used consists of the first 10,000 entries from the Pile~\cite{pile}. The throughput is measured by the number of tokens processed by each GPU per second, referred to as TGS.

As shown in Figure~\ref{fig:throughput}, on the A100 connected by NVLink, CoLLiE's throughput significantly surpasses the baseline attributed to the integration of FlashAttention. On the RTX-3090, where communication is limited by PCIe, CoLLiE achieves substantially higher throughput by a more appropriate parallelism approach, namely TP and PP.

\subsection{Empirical Assessment of Effectiveness}

\begin{table}[t]
\fontsize{8.7pt}{10pt}\selectfont
\centering
    \setlength\tabcolsep{1.3pt}
\begin{tabular}{lcccccc}
\toprule
     & MMLU & BBH  & GSM  & HumanEval & AlpacaFarm & Avg.  \\
\midrule
Vanilla   & 62.4 & 56.9 & 53.9 & 20.7      & \ \ 4.7       & 39.7 \\
LoRA & 62.7 & 58.7 & \textbf{60.5} & \textbf{32.9}      & 69.6       & \textbf{56.9} \\
LOMO & 62.1 & 56.9 & 57.6 & 28.1      & 65.2       & 54.0 \\
AdaLomo & 62.7 & \textbf{59.0} & 59.7 & 29.9      & \textbf{73.4}       & \textbf{56.9} \\
Lion & 58.2 & 52.6 & 41.3 & 25.0      & 66.2       & 48.7 \\
Adan & 57.3 & 51.9 & 37.3 & 21.3      & 62.5       & 46.1 \\
Adam & \textbf{63.0} & 58.0 & 55.3 & 28.1      & 73.1      & 55.5 \\
\bottomrule
\end{tabular}
\caption{Comparison of different training methods on GPT4-Alpaca. Instruction-tuning significantly enhances the instruction-following ability of vanilla LLaMA-65B.}
\label{tab:assess}
\end{table}
\label{sec:assessment}

We employ various training methods on GPT-4-Alpaca~\cite{gpt4-alpaca} for the LLaMA-65B model and evaluate the factual knowledge, reasonning abilities, code capabilities, and instruction-following abilities using MMLU~\cite{mmlu}, BBH~\cite{bbh}, GSM8K~\cite{gsm8k}, HumanEval~\cite{humaneval}, and AlpacaFarm~\cite{alpaca-farm}. The hyperparameters and templates for training and evaluating can be found in Appendix~\ref{appendix:hyperparameters-training} and Appendix~\ref{appendix:templates}, respectively.

The results in Table~\ref{tab:assess} demonstrate that while the vanilla LLaMA-65B already exhibits substantial capabilities, it struggles to effectively follow instructions from actual users. The performance of the models significantly improves on average after instruction-tuning. Training methods such as LoRA, LOMO, AdaLomo and AdamW significantly enhance the model's ability to follow instructions without compromising its other performance.


\section{Conclusion}
We have introduced CoLLiE, a library for collaboratively training large language models
in an efficient way.
CoLLiE offers efficient models with FlashAttention and structurally supportive for 3D parallelism. Moreover, CoLLiE provides a comprehensive and customizable \texttt{Trainer} to assist users throughout the training process, supporting various training methods. 
We have tested the relationship between the GPU memory requirements and model parameter sizes as a reference for users. In terms of throughput, CoLLiE is significantly more efficient than HuggingFace's parallel solutions. The effectiveness of different training methods are also empirically assessed on instruction-tuning tasks. 


\section*{Limitations}
We discuss the limitations of this paper from the following two aspects:

1) Although we profile the memory usage under real training conditions in this paper, a more fine-grained memory allocation situation is not provided. In the future, we plan to develop a fine-grained memory monitor to assist users in training.

2) Due to resource and time constraints, this paper only presents the instruction-tuning results of LLaMA-65B with different training methods. This restricts users from comparing the performance of models of different sizes. We will provide performances of more models under various training methods and continuously update them on our Github repository for user reference. Furthermore, while CoLLiE has implemented the Sophia optimizer to enhance pre-training efficiency, we have not conducted extensive experiments under costly pre-training tasks.





\section*{Acknowledgements}
This work was supported by the National Key Research and Development Program of China (No.2022ZD0160102) and National Natural Science Foundation of China (No.62022027). 

The CoLLiE library would also like to extend its gratitude to the early users, whose invaluable suggestions have greatly contributed to this project. Special thanks to Peijv Liu, Zhangyue Yin, Qinghui Gao, Shuowen Zhang, Qiong Tang and Junliang He.

\bibliography{anthology,custom}

\begin{thebibliography}{38}
\expandafter\ifx\csname natexlab\endcsname\relax\def\natexlab#1{#1}\fi

\bibitem[{Bian et~al.(2021)Bian, Liu, Wang, Huang, Li, Wang, Cui, and
  You}]{colossalai}
Zhengda Bian, Hongxin Liu, Boxiang Wang, Haichen Huang, Yongbin Li, Chuanrui
  Wang, Fan Cui, and Yang You. 2021.
\newblock \href {http://arxiv.org/abs/2110.14883} {Colossal-ai: {A} unified
  deep learning system for large-scale parallel training}.
\newblock \emph{CoRR}, abs/2110.14883.

\bibitem[{Chen et~al.(2021)Chen, Tworek, Jun, Yuan, de~Oliveira~Pinto, Kaplan,
  Edwards, Burda, Joseph, Brockman, Ray, Puri, Krueger, Petrov, Khlaaf, Sastry,
  Mishkin, Chan, Gray, Ryder, Pavlov, Power, Kaiser, Bavarian, Winter, Tillet,
  Such, Cummings, Plappert, Chantzis, Barnes, Herbert{-}Voss, Guss, Nichol,
  Paino, Tezak, Tang, Babuschkin, Balaji, Jain, Saunders, Hesse, Carr, Leike,
  Achiam, Misra, Morikawa, Radford, Knight, Brundage, Murati, Mayer, Welinder,
  McGrew, Amodei, McCandlish, Sutskever, and Zaremba}]{humaneval}
Mark Chen, Jerry Tworek, Heewoo Jun, Qiming Yuan, Henrique~Pond{\'{e}}
  de~Oliveira~Pinto, Jared Kaplan, Harrison Edwards, Yuri Burda, Nicholas
  Joseph, Greg Brockman, Alex Ray, Raul Puri, Gretchen Krueger, Michael Petrov,
  Heidy Khlaaf, Girish Sastry, Pamela Mishkin, Brooke Chan, Scott Gray, Nick
  Ryder, Mikhail Pavlov, Alethea Power, Lukasz Kaiser, Mohammad Bavarian,
  Clemens Winter, Philippe Tillet, Felipe~Petroski Such, Dave Cummings,
  Matthias Plappert, Fotios Chantzis, Elizabeth Barnes, Ariel Herbert{-}Voss,
  William~Hebgen Guss, Alex Nichol, Alex Paino, Nikolas Tezak, Jie Tang, Igor
  Babuschkin, Suchir Balaji, Shantanu Jain, William Saunders, Christopher
  Hesse, Andrew~N. Carr, Jan Leike, Joshua Achiam, Vedant Misra, Evan Morikawa,
  Alec Radford, Matthew Knight, Miles Brundage, Mira Murati, Katie Mayer, Peter
  Welinder, Bob McGrew, Dario Amodei, Sam McCandlish, Ilya Sutskever, and
  Wojciech Zaremba. 2021.
\newblock \href {http://arxiv.org/abs/2107.03374} {Evaluating large language
  models trained on code}.
\newblock \emph{CoRR}, abs/2107.03374.

\bibitem[{Chen et~al.(2023)Chen, Liang, Huang, Real, Wang, Liu, Pham, Dong,
  Luong, Hsieh, Lu, and Le}]{lion}
Xiangning Chen, Chen Liang, Da~Huang, Esteban Real, Kaiyuan Wang, Yao Liu, Hieu
  Pham, Xuanyi Dong, Thang Luong, Cho{-}Jui Hsieh, Yifeng Lu, and Quoc~V. Le.
  2023.
\newblock \href {https://doi.org/10.48550/arXiv.2302.06675} {Symbolic discovery
  of optimization algorithms}.
\newblock \emph{CoRR}, abs/2302.06675.

\bibitem[{Cobbe et~al.(2021)Cobbe, Kosaraju, Bavarian, Chen, Jun, Kaiser,
  Plappert, Tworek, Hilton, Nakano, Hesse, and Schulman}]{gsm8k}
Karl Cobbe, Vineet Kosaraju, Mohammad Bavarian, Mark Chen, Heewoo Jun, Lukasz
  Kaiser, Matthias Plappert, Jerry Tworek, Jacob Hilton, Reiichiro Nakano,
  Christopher Hesse, and John Schulman. 2021.
\newblock \href {http://arxiv.org/abs/2110.14168} {Training verifiers to solve
  math word problems}.
\newblock \emph{CoRR}, abs/2110.14168.

\bibitem[{Dao(2023)}]{flashattn2}
Tri Dao. 2023.
\newblock \href {https://doi.org/10.48550/arXiv.2307.08691} {Flashattention-2:
  Faster attention with better parallelism and work partitioning}.
\newblock \emph{CoRR}, abs/2307.08691.

\bibitem[{Dao et~al.(2022)Dao, Fu, Ermon, Rudra, and R{\'{e}}}]{flashattn}
Tri Dao, Daniel~Y. Fu, Stefano Ermon, Atri Rudra, and Christopher R{\'{e}}.
  2022.
\newblock \href
  {http://papers.nips.cc/paper\_files/paper/2022/hash/67d57c32e20fd0a7a302cb81d36e40d5-Abstract-Conference.html}
  {Flashattention: Fast and memory-efficient exact attention with
  io-awareness}.
\newblock In \emph{NeurIPS}.

\bibitem[{Diao et~al.(2023)Diao, Pan, Dong, Shum, Zhang, Xiong, and
  Zhang}]{lmflow}
Shizhe Diao, Rui Pan, Hanze Dong, Kashun Shum, Jipeng Zhang, Wei Xiong, and
  Tong Zhang. 2023.
\newblock \href {https://doi.org/10.48550/arXiv.2306.12420} {Lmflow: An
  extensible toolkit for finetuning and inference of large foundation models}.
\newblock \emph{CoRR}, abs/2306.12420.

\bibitem[{Ding et~al.(2023)Ding, Qin, Yang, Wei, Yang, Su, Hu, Chen, Chan,
  Chen, Yi, Zhao, Wang, Liu, Zheng, Chen, Liu, Tang, Li, and
  Sun}]{peft_overview}
Ning Ding, Yujia Qin, Guang Yang, Fuchao Wei, Zonghan Yang, Yusheng Su,
  Shengding Hu, Yulin Chen, Chi{-}Min Chan, Weize Chen, Jing Yi, Weilin Zhao,
  Xiaozhi Wang, Zhiyuan Liu, Hai{-}Tao Zheng, Jianfei Chen, Yang Liu, Jie Tang,
  Juanzi Li, and Maosong Sun. 2023.
\newblock \href {https://doi.org/10.1038/s42256-023-00626-4}
  {Parameter-efficient fine-tuning of large-scale pre-trained language models}.
\newblock \emph{Nat. Mac. Intell.}, 5(3):220--235.

\bibitem[{Du et~al.(2022)Du, Qian, Liu, Ding, Qiu, Yang, and Tang}]{glm}
Zhengxiao Du, Yujie Qian, Xiao Liu, Ming Ding, Jiezhong Qiu, Zhilin Yang, and
  Jie Tang. 2022.
\newblock Glm: General language model pretraining with autoregressive blank
  infilling.
\newblock In \emph{Proceedings of the 60th Annual Meeting of the Association
  for Computational Linguistics (Volume 1: Long Papers)}, pages 320--335.

\bibitem[{Dubois et~al.(2023)Dubois, Li, Taori, Zhang, Gulrajani, Ba, Guestrin,
  Liang, and Hashimoto}]{alpaca-farm}
Yann Dubois, Xuechen Li, Rohan Taori, Tianyi Zhang, Ishaan Gulrajani, Jimmy Ba,
  Carlos Guestrin, Percy Liang, and Tatsunori~B. Hashimoto. 2023.
\newblock \href {https://doi.org/10.48550/arXiv.2305.14387} {Alpacafarm: {A}
  simulation framework for methods that learn from human feedback}.
\newblock \emph{CoRR}, abs/2305.14387.

\bibitem[{Gao et~al.(2021)Gao, Biderman, Black, Golding, Hoppe, Foster, Phang,
  He, Thite, Nabeshima, Presser, and Leahy}]{pile}
Leo Gao, Stella Biderman, Sid Black, Laurence Golding, Travis Hoppe, Charles
  Foster, Jason Phang, Horace He, Anish Thite, Noa Nabeshima, Shawn Presser,
  and Connor Leahy. 2021.
\newblock \href {http://arxiv.org/abs/2101.00027} {The pile: An 800gb dataset
  of diverse text for language modeling}.
\newblock \emph{CoRR}, abs/2101.00027.

\bibitem[{Hendrycks et~al.(2021)Hendrycks, Burns, Basart, Zou, Mazeika, Song,
  and Steinhardt}]{mmlu}
Dan Hendrycks, Collin Burns, Steven Basart, Andy Zou, Mantas Mazeika, Dawn
  Song, and Jacob Steinhardt. 2021.
\newblock \href {https://openreview.net/forum?id=d7KBjmI3GmQ} {Measuring
  massive multitask language understanding}.
\newblock In \emph{9th International Conference on Learning Representations,
  {ICLR} 2021, Virtual Event, Austria, May 3-7, 2021}. OpenReview.net.

\bibitem[{Houlsby et~al.(2019)Houlsby, Giurgiu, Jastrzebski, Morrone,
  de~Laroussilhe, Gesmundo, Attariyan, and Gelly}]{adapter-tuning}
Neil Houlsby, Andrei Giurgiu, Stanislaw Jastrzebski, Bruna Morrone, Quentin
  de~Laroussilhe, Andrea Gesmundo, Mona Attariyan, and Sylvain Gelly. 2019.
\newblock \href {http://proceedings.mlr.press/v97/houlsby19a.html}
  {Parameter-efficient transfer learning for {NLP}}.
\newblock In \emph{Proceedings of the 36th International Conference on Machine
  Learning, {ICML} 2019, 9-15 June 2019, Long Beach, California, {USA}},
  volume~97 of \emph{Proceedings of Machine Learning Research}, pages
  2790--2799. {PMLR}.

\bibitem[{Hu et~al.(2022)Hu, Shen, Wallis, Allen{-}Zhu, Li, Wang, Wang, and
  Chen}]{lora}
Edward~J. Hu, Yelong Shen, Phillip Wallis, Zeyuan Allen{-}Zhu, Yuanzhi Li,
  Shean Wang, Lu~Wang, and Weizhu Chen. 2022.
\newblock \href {https://openreview.net/forum?id=nZeVKeeFYf9} {Lora: Low-rank
  adaptation of large language models}.
\newblock In \emph{The Tenth International Conference on Learning
  Representations, {ICLR} 2022, Virtual Event, April 25-29, 2022}.
  OpenReview.net.

\bibitem[{Huang et~al.(2019)Huang, Cheng, Bapna, Firat, Chen, Chen, Lee, Ngiam,
  Le, Wu, and Chen}]{gpipe}
Yanping Huang, Youlong Cheng, Ankur Bapna, Orhan Firat, Dehao Chen, Mia~Xu
  Chen, HyoukJoong Lee, Jiquan Ngiam, Quoc~V. Le, Yonghui Wu, and Zhifeng Chen.
  2019.
\newblock \href
  {https://proceedings.neurips.cc/paper/2019/hash/093f65e080a295f8076b1c5722a46aa2-Abstract.html}
  {Gpipe: Efficient training of giant neural networks using pipeline
  parallelism}.
\newblock In \emph{Advances in Neural Information Processing Systems 32: Annual
  Conference on Neural Information Processing Systems 2019, NeurIPS 2019,
  December 8-14, 2019, Vancouver, BC, Canada}, pages 103--112.

\bibitem[{Kingma and Ba(2015)}]{adam}
Diederik~P. Kingma and Jimmy Ba. 2015.
\newblock \href {http://arxiv.org/abs/1412.6980} {Adam: {A} method for
  stochastic optimization}.
\newblock In \emph{3rd International Conference on Learning Representations,
  {ICLR} 2015, San Diego, CA, USA, May 7-9, 2015, Conference Track
  Proceedings}.

\bibitem[{Lester et~al.(2021)Lester, Al{-}Rfou, and Constant}]{prompt-tuning}
Brian Lester, Rami Al{-}Rfou, and Noah Constant. 2021.
\newblock \href {https://doi.org/10.18653/v1/2021.emnlp-main.243} {The power of
  scale for parameter-efficient prompt tuning}.
\newblock In \emph{Proceedings of the 2021 Conference on Empirical Methods in
  Natural Language Processing, {EMNLP} 2021, Virtual Event / Punta Cana,
  Dominican Republic, 7-11 November, 2021}, pages 3045--3059. Association for
  Computational Linguistics.

\bibitem[{Li and Liang(2021)}]{prefix-tuning}
Xiang~Lisa Li and Percy Liang. 2021.
\newblock \href {https://doi.org/10.18653/v1/2021.acl-long.353} {Prefix-tuning:
  Optimizing continuous prompts for generation}.
\newblock In \emph{Proceedings of the 59th Annual Meeting of the Association
  for Computational Linguistics and the 11th International Joint Conference on
  Natural Language Processing, {ACL/IJCNLP} 2021, (Volume 1: Long Papers),
  Virtual Event, August 1-6, 2021}, pages 4582--4597. Association for
  Computational Linguistics.

\bibitem[{Liu et~al.(2023)Liu, Li, Hall, Liang, and Ma}]{sophia}
Hong Liu, Zhiyuan Li, David Hall, Percy Liang, and Tengyu Ma. 2023.
\newblock \href {https://doi.org/10.48550/arXiv.2305.14342} {Sophia: {A}
  scalable stochastic second-order optimizer for language model pre-training}.
\newblock \emph{CoRR}, abs/2305.14342.

\bibitem[{Lv et~al.(2023{\natexlab{a}})Lv, Yan, Guo, Lv, and Qiu}]{adalomo}
Kai Lv, Hang Yan, Qipeng Guo, Haijun Lv, and Xipeng Qiu. 2023{\natexlab{a}}.
\newblock \href {https://doi.org/10.48550/ARXIV.2310.10195} {Adalomo:
  Low-memory optimization with adaptive learning rate}.
\newblock \emph{CoRR}, abs/2310.10195.

\bibitem[{Lv et~al.(2023{\natexlab{b}})Lv, Yang, Liu, Gao, Guo, and Qiu}]{lomo}
Kai Lv, Yuqing Yang, Tengxiao Liu, Qinghui Gao, Qipeng Guo, and Xipeng Qiu.
  2023{\natexlab{b}}.
\newblock \href {https://doi.org/10.48550/arXiv.2306.09782} {Full parameter
  fine-tuning for large language models with limited resources}.
\newblock \emph{CoRR}, abs/2306.09782.

\bibitem[{Mangrulkar et~al.(2022)Mangrulkar, Gugger, Debut, Belkada, and
  Paul}]{peft}
Sourab Mangrulkar, Sylvain Gugger, Lysandre Debut, Younes Belkada, and Sayak
  Paul. 2022.
\newblock Peft: State-of-the-art parameter-efficient fine-tuning methods.
\newblock \url{https://github.com/huggingface/peft}.

\bibitem[{Narayanan et~al.(2019)Narayanan, Harlap, Phanishayee, Seshadri,
  Devanur, Ganger, Gibbons, and Zaharia}]{pipedream}
Deepak Narayanan, Aaron Harlap, Amar Phanishayee, Vivek Seshadri, Nikhil~R.
  Devanur, Gregory~R. Ganger, Phillip~B. Gibbons, and Matei Zaharia. 2019.
\newblock \href {https://doi.org/10.1145/3341301.3359646} {Pipedream:
  generalized pipeline parallelism for {DNN} training}.
\newblock In \emph{Proceedings of the 27th {ACM} Symposium on Operating Systems
  Principles, {SOSP} 2019, Huntsville, ON, Canada, October 27-30, 2019}, pages
  1--15. {ACM}.

\bibitem[{Paszke et~al.(2019)Paszke, Gross, Massa, Lerer, Bradbury, Chanan,
  Killeen, Lin, Gimelshein, Antiga, Desmaison, K{\"{o}}pf, Yang, DeVito,
  Raison, Tejani, Chilamkurthy, Steiner, Fang, Bai, and Chintala}]{pytorch}
Adam Paszke, Sam Gross, Francisco Massa, Adam Lerer, James Bradbury, Gregory
  Chanan, Trevor Killeen, Zeming Lin, Natalia Gimelshein, Luca Antiga, Alban
  Desmaison, Andreas K{\"{o}}pf, Edward~Z. Yang, Zachary DeVito, Martin Raison,
  Alykhan Tejani, Sasank Chilamkurthy, Benoit Steiner, Lu~Fang, Junjie Bai, and
  Soumith Chintala. 2019.
\newblock \href
  {https://proceedings.neurips.cc/paper/2019/hash/bdbca288fee7f92f2bfa9f7012727740-Abstract.html}
  {Pytorch: An imperative style, high-performance deep learning library}.
\newblock In \emph{Advances in Neural Information Processing Systems 32: Annual
  Conference on Neural Information Processing Systems 2019, NeurIPS 2019,
  December 8-14, 2019, Vancouver, BC, Canada}, pages 8024--8035.

\bibitem[{Peng et~al.(2023)Peng, Li, He, Galley, and Gao}]{gpt4-alpaca}
Baolin Peng, Chunyuan Li, Pengcheng He, Michel Galley, and Jianfeng Gao. 2023.
\newblock \href {https://doi.org/10.48550/arXiv.2304.03277} {Instruction tuning
  with {GPT-4}}.
\newblock \emph{CoRR}, abs/2304.03277.

\bibitem[{Rajbhandari et~al.(2020)Rajbhandari, Rasley, Ruwase, and He}]{zero}
Samyam Rajbhandari, Jeff Rasley, Olatunji Ruwase, and Yuxiong He. 2020.
\newblock \href {https://doi.org/10.1109/SC41405.2020.00024} {Zero: memory
  optimizations toward training trillion parameter models}.
\newblock In \emph{Proceedings of the International Conference for High
  Performance Computing, Networking, Storage and Analysis, {SC} 2020, Virtual
  Event / Atlanta, Georgia, USA, November 9-19, 2020}, page~20. {IEEE/ACM}.

\bibitem[{Rasley et~al.(2020)Rasley, Rajbhandari, Ruwase, and He}]{deepspeed}
Jeff Rasley, Samyam Rajbhandari, Olatunji Ruwase, and Yuxiong He. 2020.
\newblock \href {https://doi.org/10.1145/3394486.3406703} {Deepspeed: System
  optimizations enable training deep learning models with over 100 billion
  parameters}.
\newblock In \emph{{KDD} '20: The 26th {ACM} {SIGKDD} Conference on Knowledge
  Discovery and Data Mining, Virtual Event, CA, USA, August 23-27, 2020}, pages
  3505--3506. {ACM}.

\bibitem[{Scao et~al.(2022)Scao, Fan, Akiki, Pavlick, Ilic, Hesslow,
  Castagn{\'{e}}, Luccioni, Yvon, Gall{\'{e}}, Tow, Rush, Biderman, Webson,
  Ammanamanchi, Wang, Sagot, Muennighoff, del Moral, Ruwase, Bawden, Bekman,
  McMillan{-}Major, Beltagy, Nguyen, Saulnier, Tan, Suarez, Sanh,
  Lauren{\c{c}}on, Jernite, Launay, Mitchell, Raffel, Gokaslan, Simhi, Soroa,
  Aji, Alfassy, Rogers, Nitzav, Xu, Mou, Emezue, Klamm, Leong, van Strien,
  Adelani, and et~al.}]{bloom}
Teven~Le Scao, Angela Fan, Christopher Akiki, Ellie Pavlick, Suzana Ilic,
  Daniel Hesslow, Roman Castagn{\'{e}}, Alexandra~Sasha Luccioni,
  Fran{\c{c}}ois Yvon, Matthias Gall{\'{e}}, Jonathan Tow, Alexander~M. Rush,
  Stella Biderman, Albert Webson, Pawan~Sasanka Ammanamanchi, Thomas Wang,
  Beno{\^{\i}}t Sagot, Niklas Muennighoff, Albert~Villanova del Moral, Olatunji
  Ruwase, Rachel Bawden, Stas Bekman, Angelina McMillan{-}Major, Iz~Beltagy,
  Huu Nguyen, Lucile Saulnier, Samson Tan, Pedro~Ortiz Suarez, Victor Sanh,
  Hugo Lauren{\c{c}}on, Yacine Jernite, Julien Launay, Margaret Mitchell, Colin
  Raffel, Aaron Gokaslan, Adi Simhi, Aitor Soroa, Alham~Fikri Aji, Amit
  Alfassy, Anna Rogers, Ariel~Kreisberg Nitzav, Canwen Xu, Chenghao Mou, Chris
  Emezue, Christopher Klamm, Colin Leong, Daniel van Strien, David~Ifeoluwa
  Adelani, and et~al. 2022.
\newblock \href {https://doi.org/10.48550/arXiv.2211.05100} {{BLOOM:} {A}
  176b-parameter open-access multilingual language model}.
\newblock \emph{CoRR}, abs/2211.05100.

\bibitem[{Shoeybi et~al.(2019)Shoeybi, Patwary, Puri, LeGresley, Casper, and
  Catanzaro}]{megatron}
Mohammad Shoeybi, Mostofa Patwary, Raul Puri, Patrick LeGresley, Jared Casper,
  and Bryan Catanzaro. 2019.
\newblock \href {http://arxiv.org/abs/1909.08053} {Megatron-lm: Training
  multi-billion parameter language models using model parallelism}.
\newblock \emph{CoRR}, abs/1909.08053.

\bibitem[{Sun et~al.(2023{\natexlab{a}})Sun, Zhang, He, Li, Cheng, Yan, Liu,
  Shao, Tang, Zhao, Chen, Zheng, Zhou, Li, Zhan, Zhou, Li, Yang, Wu, Yin,
  Huang, and Qiu}]{moss}
Tianxiang Sun, Xiaotian Zhang, Zhengfu He, Peng Li, Qinyuan Cheng, Hang Yan,
  Xiangyang Liu, Yunfan Shao, Qiong Tang, Xingjian Zhao, Ke~Chen, Yining Zheng,
  Zhejian Zhou, Ruixiao Li, Jun Zhan, Yunhua Zhou, Linyang Li, Xiaogui Yang,
  Lingling Wu, Zhangyue Yin, Xuanjing Huang, and Xipeng Qiu.
  2023{\natexlab{a}}.
\newblock Moss: Training conversational language models from synthetic data.
\newblock \url{https://github.com/OpenLMLab/MOSS}.

\bibitem[{Sun et~al.(2023{\natexlab{b}})Sun, Ji, Ma, and
  Li}]{compare_peft_full-tuning}
Xianghui Sun, Yunjie Ji, Baochang Ma, and Xiangang Li. 2023{\natexlab{b}}.
\newblock \href {https://doi.org/10.48550/arXiv.2304.08109} {A comparative
  study between full-parameter and lora-based fine-tuning on chinese
  instruction data for instruction following large language model}.
\newblock \emph{CoRR}, abs/2304.08109.

\bibitem[{Suzgun et~al.(2023)Suzgun, Scales, Sch{\"{a}}rli, Gehrmann, Tay,
  Chung, Chowdhery, Le, Chi, Zhou, and Wei}]{bbh}
Mirac Suzgun, Nathan Scales, Nathanael Sch{\"{a}}rli, Sebastian Gehrmann,
  Yi~Tay, Hyung~Won Chung, Aakanksha Chowdhery, Quoc~V. Le, Ed~Chi, Denny Zhou,
  and Jason Wei. 2023.
\newblock \href {https://aclanthology.org/2023.findings-acl.824} {Challenging
  big-bench tasks and whether chain-of-thought can solve them}.
\newblock In \emph{Findings of the Association for Computational Linguistics:
  {ACL} 2023, Toronto, Canada, July 9-14, 2023}, pages 13003--13051.
  Association for Computational Linguistics.

\bibitem[{Team(2023)}]{internlm}
InternLM Team. 2023.
\newblock Internlm: A multilingual language model with progressively enhanced
  capabilities.
\newblock \url{https://github.com/InternLM/InternLM}.

\bibitem[{Touvron et~al.(2023)Touvron, Lavril, Izacard, Martinet, Lachaux,
  Lacroix, Rozi{\`{e}}re, Goyal, Hambro, Azhar, Rodriguez, Joulin, Grave, and
  Lample}]{llama}
Hugo Touvron, Thibaut Lavril, Gautier Izacard, Xavier Martinet, Marie{-}Anne
  Lachaux, Timoth{\'{e}}e Lacroix, Baptiste Rozi{\`{e}}re, Naman Goyal, Eric
  Hambro, Faisal Azhar, Aur{\'{e}}lien Rodriguez, Armand Joulin, Edouard Grave,
  and Guillaume Lample. 2023.
\newblock \href {https://doi.org/10.48550/arXiv.2302.13971} {Llama: Open and
  efficient foundation language models}.
\newblock \emph{CoRR}, abs/2302.13971.

\bibitem[{Wang et~al.(2023)Wang, Ivison, Dasigi, Hessel, Khot, Chandu, Wadden,
  MacMillan, Smith, Beltagy, and Hajishirzi}]{tulu}
Yizhong Wang, Hamish Ivison, Pradeep Dasigi, Jack Hessel, Tushar Khot,
  Khyathi~Raghavi Chandu, David Wadden, Kelsey MacMillan, Noah~A. Smith,
  Iz~Beltagy, and Hannaneh Hajishirzi. 2023.
\newblock \href {https://doi.org/10.48550/arXiv.2306.04751} {How far can camels
  go? exploring the state of instruction tuning on open resources}.
\newblock \emph{CoRR}, abs/2306.04751.

\bibitem[{Wolf et~al.(2020)Wolf, Debut, Sanh, Chaumond, Delangue, Moi, Cistac,
  Rault, Louf, Funtowicz, Davison, Shleifer, von Platen, Ma, Jernite, Plu, Xu,
  Scao, Gugger, Drame, Lhoest, and Rush}]{transformers}
Thomas Wolf, Lysandre Debut, Victor Sanh, Julien Chaumond, Clement Delangue,
  Anthony Moi, Pierric Cistac, Tim Rault, Rémi Louf, Morgan Funtowicz, Joe
  Davison, Sam Shleifer, Patrick von Platen, Clara Ma, Yacine Jernite, Julien
  Plu, Canwen Xu, Teven~Le Scao, Sylvain Gugger, Mariama Drame, Quentin Lhoest,
  and Alexander~M. Rush. 2020.
\newblock \href {https://www.aclweb.org/anthology/2020.emnlp-demos.6}
  {Transformers: State-of-the-art natural language processing}.
\newblock In \emph{Proceedings of the 2020 Conference on Empirical Methods in
  Natural Language Processing: System Demonstrations}, pages 38--45, Online.
  Association for Computational Linguistics.

\bibitem[{Xie et~al.(2022)Xie, Zhou, Li, Lin, and Yan}]{adan}
Xingyu Xie, Pan Zhou, Huan Li, Zhouchen Lin, and Shuicheng Yan. 2022.
\newblock \href {https://doi.org/10.48550/arXiv.2208.06677} {Adan: Adaptive
  nesterov momentum algorithm for faster optimizing deep models}.
\newblock \emph{CoRR}, abs/2208.06677.

\bibitem[{Zhang et~al.(2022)Zhang, Roller, Goyal, Artetxe, Chen, Chen, Dewan,
  Diab, Li, Lin, Mihaylov, Ott, Shleifer, Shuster, Simig, Koura, Sridhar, Wang,
  and Zettlemoyer}]{opt}
Susan Zhang, Stephen Roller, Naman Goyal, Mikel Artetxe, Moya Chen, Shuohui
  Chen, Christopher Dewan, Mona~T. Diab, Xian Li, Xi~Victoria Lin, Todor
  Mihaylov, Myle Ott, Sam Shleifer, Kurt Shuster, Daniel Simig, Punit~Singh
  Koura, Anjali Sridhar, Tianlu Wang, and Luke Zettlemoyer. 2022.
\newblock \href {https://doi.org/10.48550/arXiv.2205.01068} {{OPT:} open
  pre-trained transformer language models}.
\newblock \emph{CoRR}, abs/2205.01068.

\end{thebibliography}
\bibliographystyle{acl_natbib}

\clearpage
\appendix

\section{Code Example}
Listing~\ref{lst:code_example} presents the simplest code example for training with CoLLiE.
\label{sec:example}
\begin{figure*}
    \begin{lstlisting}[style=mystyle, caption={An example for training with CoLLiE.}, label={lst:code_example}]
import torch
from collie.config import CollieConfig
from collie.models import LlamaForCausalLM
from collie.controller import Trainer
model_name_or_path = 'meta-llama/Llama-2-7b-hf'
# load model config from huggingface hub
config = CollieConfig.from_pretrained(model_name_or_path)
# set pipeline parallelism size via config
config.pp_size = 8
# load pre-trained weights from huggingface hub 
# and partition the weights into 8 stages for pipeline parallelism
model = LlamaForCausalLM.from_pretrained(
    model_name_or_path,
    config=config
)
optimizer = torch.optim.AdamW(model.parameters(), lr=2e-5)
# one of the two formats collie defined for training
train_dataset = [
    {'text': 'Collie is a package for training large language models.'}
    for _ in range(100)
]
trainer = Trainer(
    model=model,
    optimizer=optimizer,
    config=config,
    train_dataset=train_dataset,
)
# start the training process
trainer.train()
\end{lstlisting}
\end{figure*}

\section{Hyperparameters}

\begin{table}[h]
    \centering
    \begin{tabular}{ccccc}
    \toprule
\begin{tabular}[c]{@{}c@{}}Training\\ Methods\end{tabular} & LR   & \begin{tabular}[c]{@{}c@{}}Batch\\ Size\end{tabular} & \begin{tabular}[c]{@{}c@{}}Weight\\ Decay\end{tabular} & Epochs \\
\midrule
LoRA                                                       & 3e-4 & 128                                                  & 1e-2                                                   & 3      \\
LOMO                                                       & 1e-2 & 16                                                   & -                                                      & 5      \\
AdaLomo                                                       & 5e-4 & 128                                                   & -                                                      & 5      \\
Lion                                                       & 3e-6 & 128                                                  & 3e-2                                                   & 3      \\
Adan                                                       & 5e-5 & 128                                                  & 2e-2                                                   & 3      \\
Adam                                                       & 1e-5 & 128                                                  & 1e-2                                                   & 3     \\
\bottomrule
\end{tabular}
    \caption{Hyperparameters for training.}
    \label{tab:hyper-it}
\end{table}

\begin{table}
\begin{tabularx}{0.48\textwidth}{X}
  \toprule
\textbf{Template for entries with input} \\
\midrule
Below is an instruction that describes a task, paired with an input that provides further context. Write a response that appropriately completes the request.\\\\
\#\#\# Instruction:\\
\{instruction\}\\\\
\#\#\# Input:\\
\{input\}\\\\
\#\#\# Response:\{response\}\\
\toprule
\textbf{Template for entries without input} \\
\midrule
Below is an instruction that describes a task. Write a response that appropriately completes the request.\\\\
\#\#\# Instruction:\\
\{instruction\}\\\\
\#\#\# Response:\{response\} \\
\bottomrule
\end{tabularx}
\caption{Templates used for training.}
\label{tab:tem_train}
\end{table}
\subsection{Memory Requirements}
We choose the combination of Tensor Parallelism (TP) and Pipeline Parallelism (PP) as our parallelism strategy. The batch size is set to 2048, and the gradient accumulation steps are set to 2. It's worth noting that increasing the value of the gradient accumulation steps would not significantly increase the memory usage.

\subsection{Throughput}
In our throughput tests, we consistently employ Adam as the optimizer. We utilize the default settings of DeepSpeed for ZeRO-3 and strive to maximize the micro batch size to enhance throughput. For Tensor Parallelism/Pipeline Parallelism (TP/PP), we ensure that the gradient accumulation steps are more than four times the number of pipeline stages to minimize the bubble. The specific configurations are illustrated in Table~\ref{tab:hyper-throughput}.

\begin{table*}
    \small
    \centering
    \begin{tabular}{cccccc}
    \toprule
    Model Params           &                         & 7B                 & 13B                & 30B                & 65B                \\
    \midrule
    Device                 & \multicolumn{5}{c}{A100}                                                                                      \\
    Mode                   & \multicolumn{5}{c}{Fine-tune / Pre-train}                                                                       \\
    \midrule
    \# GPU                 &                         & 4                  & 8                  & 16                 & 32                 \\
    \midrule
    HuggingFace with ZeRO-3 & Batch Size,\ GAS          & 64,2 / 64,16       & 64,2 / 64,16       & 64, 2 / 128,8      & 128,1 / 128,8      \\
    CoLLiE with ZeRO-3     & Batch Size,\ GAS          & 64,2 / 64,16       & 128,1 / 64,16      & 128,1 / 128,8      & 128,1 / 128,8      \\
    CoLLiE with TP/PP      & Batch Size,\ GAS          & 4,32 / 8,128       & 2,64 / 16,64       & 1,128 / 2,512      & 1,128 / 1,1024     \\
    \midrule
    Device                 & \multicolumn{5}{c}{RTX-3090}                                                                                  \\
    Mode                   & \multicolumn{5}{c}{Fine-tune / Pre-train}                                                                       \\
    \midrule
    \# GPU                 &                         & 8                  & 24                 & 48                 & -                  \\
    \midrule
    HuggingFace with ZeRO-3 & Batch Size,\ GAS          & 8,16 / 8,128       & 24,6 / 24,43       & 48,3 / 48,22       & -                  \\
    CoLLiE with ZeRO-3     & Batch Size,\ GAS          & 8,16 / 8,128       & 24,6 / 24,43       & 48,3 / 48,22       & -                  \\
    CoLLiE with TP/PP      & Batch Size,\ GAS          & 1,128 / 1,1024     & 1,128 / 1,1024     & 1,128 / 1,1024     & -                  \\
    \bottomrule
    \end{tabular}
    \caption{\textbf{Hyperparameters for testing throughput.} We report the number of model parameters (Model Params), device (Device), mode (Mode) and number of GPU (\#GPU). We also report the corresponding batch size (Batch Size) and GAS (Gradient Accumulation Steps) for HuggingFace with ZeRO-3, CoLLiE with ZeRO-3 and CoLLiE with TP/PP.}
    \label{tab:hyper-throughput}
\end{table*}

\subsection{Instruction-tuning}
\label{appendix:hyperparameters-training}

As shown in Table~\ref{tab:hyper-it}, we have adopted the learning rates and batch sizes from the Tulu~\cite{tulu} and Alpaca-LoRA projects\footnote{\href{https://github.com/tloen/alpaca-lora}{https://github.com/tloen/alpaca-lora}} for AdamW and LoRA. To achieve better performance for LoRA, we have replaced all modules with LoRA layers, not just the q-v module. For Lion and Adan, we have used the learning rates recommended in the paper. Specifically, the learning rate for Lion is 3-10 times smaller than that of AdamW, with the weight decay correspondingly 3-10 times larger. The learning rate for the Adan optimizer is 5-10 times larger than that of AdamW, with a weight decay of 0.02. For the LOMO optimizer, which is similar to SGD, we have utilized a larger learning rate and a smaller batch size. AdaLomo is suitable for a larger learning rate than AdamW, which we set to 5e-4.

\section{Templates}
\label{appendix:templates}
\subsection{Alpaca}

We follow the template provided by the Alpaca repository\footnote{\href{https://github.com/tatsu-lab/stanford\_alpaca}{https://github.com/tatsu-lab/stanford\_alpaca}} for training, as shown in Table~\ref{tab:tem_train}.

\subsection{Evaluation}
We modify the evaluation template based on the template used during training, as shown in Table~\ref{tab:temp_eval}. The template used for evaluate on AlpacaFarm is identical to that of training on Alpaca.
\onecolumn
\begin{center}
\begin{small}
\begin{xtabular*}{0.95\textwidth}{p{0.92\textwidth}}
\toprule
\textbf{MMLU} \\
\midrule
Below is an instruction that describes a task, paired with an input that provides further context. Write a response that appropriately completes the request.\\\\\#\#\# Instruction:\\The following is a multiple choice question (with answers) about abstract algebra. You need to answer the question by selecting the correct option.\\\\\#\#\# Input:\\Find all c in Z\_3 such that Z\_3[x]/(x\^{} 2 + c) is a field.\\A. 0\\B. 1\\C. 2\\D. 3\\\\\#\#\# Response: B\\\\
Below is an instruction that describes a task, paired with an input that provides further context. Write a response that appropriately completes the request.\\\\\#\#\# Instruction:\\The following is multiple choice question (with answers) about abstract algebra. You need to answer the question by selecting the correct option.\\\\\#\#\# Input:\\Statement 1 | If aH is an element of a factor group, then |aH| divides |a|. Statement 2 | If H and K are subgroups of G then HK is a subgroup of G.\\A. True, True\\B. False, False\\C. True, False\\D. False, True\\\\\#\#\# Response: B\\\\
Below is an instruction that describes a task, paired with an input that provides further context. Write a response that appropriately completes the request.\\\\\#\#\# Instruction:\\The following is multiple choice question (with answers) about abstract algebra. You need to answer the question by selecting the correct option.\\\\\#\#\# Input:\\Statement 1 | Every element of a group generates a cyclic subgroup of the group. Statement 2 | The symmetric group S\_10 has 10 elements.\\A. True, True\\B. False, False\\C. True, False\\D. False, True\\\\\#\#\# Response: C\\\\
Below is an instruction that describes a task, paired with an input that provides further context. Write a response that appropriately completes the request.\\\\\#\#\# Instruction:\\The following is multiple choice question (with answers) about abstract algebra. You need to answer the question by selecting the correct option.\\\\\#\#\# Input:\\Statement 1| Every function from a finite set onto itself must be one to one. Statement 2 | Every subgroup of an abelian group is abelian.\\A. True, True\\B. False, False\\C. True, False\\D. False, True\\\\\#\#\# Response: A\\\\
Below is an instruction that describes a task, paired with an input that provides further context. Write a response that appropriately completes the request.\\\\\#\#\# Instruction:\\The following is multiple choice question (with answers) about abstract algebra. You need to answer the question by selecting the correct option.\\\\\#\#\# Input:\\Find the characteristic of the ring 2Z.\\A. 0\\B. 3\\C. 12\\D. 30\\\\\#\#\# Response: A\\\\
Below is an instruction that describes a task, paired with an input that provides further context. Write a response that appropriately completes the request.\\\\\#\#\# Instruction:\\The following is multiple choice question (with answers) about abstract algebra. You need to answer the question by selecting the correct option.\\\\\#\#\# Input:\\{Input}\\\\\#\#\# Response:\\
\toprule
\textbf{BBH} \\
\midrule
Below is an instruction that describes a task, paired with an input that provides further context. Write a response that appropriately completes the request.\\\\\#\#\# Instruction:\\Evaluate the result of a random Boolean expression.\\\\\#\#\# Input:\\not ( ( not not True ) ) is\\\\\#\#\# Response: Let's think step by step.\\Remember that (i) expressions inside brackets are always evaluated first and that (ii) the order of operations from highest priority to lowest priority is "not", "and", "or", respectively.\\We first simplify this expression "Z" as follows: "Z = not ( ( not not True ) ) = not ( ( A ) )" where "A = not not True".\\Let's evaluate A: A = not not True = not (not True) = not False = True.\\Plugging in A, we get: Z = not ( ( A ) ) = not ( ( True ) ) = not True = False. So the answer is False.\\
Below is an instruction that describes a task, paired with an input that provides further context. Write a response that appropriately completes the request.\\\\\#\#\# Instruction:\\Evaluate the result of a random Boolean expression.\\\\\#\#\# Input:\\True and False and not True and True is\\\\\#\#\# Response: Let's think step by step.\\Remember that (i) expressions inside brackets are always evaluated first and that (ii) the order of operations from highest priority to lowest priority is "not", "and", "or", respectively.\\We first simplify this expression "Z" as follows: "Z = True and False and not True and True = A and B" where "A = True and False" and "B = not True and True".\\Let's evaluate A: A = True and False = False.\\Let's evaluate B: B = not True and True = not (True and True) = not (True) = False.\\Plugging in A and B, we get: Z = A and B = False and False = False. So the answer is False.\\
Below is an instruction that describes a task, paired with an input that provides further context. Write a response that appropriately completes the request.\\\\\#\#\# Instruction:\\Evaluate the result of a random Boolean expression.\\\\\#\#\# Input:\\not not ( not ( False ) ) is\\\#\#\# Response: Let's think step by step.\\Remember that (i) expressions inside brackets are always evaluated first and that (ii) the order of operations from highest priority to lowest priority is "not", "and", "or", respectively.\\We first simplify this expression "Z" as follows: "Z = not not ( not ( False ) ) = not not ( A )" where "A = not ( False )".\\Let's evaluate A: A = not ( False ) = not False = True.\\Plugging in A, we get: Z = not not ( A ) = not not (True) = not not False = True. So the answer is True.\\
Below is an instruction that describes a task, paired with an input that provides further context. Write a response that appropriately completes the request.\\\#\#\# Instruction:\\Evaluate the result of a random Boolean expression.\\\\\#\#\# Input:\\\\\{input\}\\\\\#\#\# Response: Let's think step by step.\\Remember that (i) expressions inside brackets are always evaluated first and that (ii) the order of operations from highest priority to lowest priority is "not", "and", "or", respectively.\\
\toprule
\textbf{GSM8K} \\
\midrule
Below is an instruction that describes a task, paired with an input that provides further context. Write a response that appropriately completes the request.\\\\\#\#\# 
Instruction:\\Given a problem scenario with numerical data, perform the necessary calculations and provide a detailed step-by-step solution, ending the response with 'The answer is'.\\\\\#\#\# Input:\\Angelo and Melanie want to plan how many hours over the next week they should study together for their test next week. They have 2 chapters of their textbook to study and 4 worksheets to memorize. They figure out that they should dedicate 3 hours to each chapter of their textbook and 1.5 hours for each worksheet. If they plan to study no more than 4 hours each day, how many days should they plan to study total over the next week if they take a 10-minute break every hour, include 3 10-minute snack breaks each day, and 30 minutes for lunch each day?
\\\#\#\# Response:
Let's think step by step. Angelo and Melanie think they should dedicate 3 hours to each of the 2 chapters, 3 hours x 2 chapters = 6 hours total.
For the worksheets they plan to dedicate 1.5 hours for each worksheet, 1.5 hours x 4 worksheets = 6 hours total.
Angelo and Melanie need to start with planning 12 hours to study, at 4 hours a day, 12 / 4 = 3 days.
However, they need to include time for breaks and lunch. Every hour they want to include a 10-minute break, so 12 total hours x 10 minutes = 120 extra minutes for breaks.
They also want to include 3 10-minute snack breaks, 3 x 10 minutes = 30 minutes.
And they want to include 30 minutes for lunch each day, so 120 minutes for breaks + 30 minutes for snack breaks + 30 minutes for lunch = 180 minutes, or 180 / 60 minutes per hour = 3 extra hours.
So Angelo and Melanie want to plan 12 hours to study + 3 hours of breaks = 15 hours total.
They want to study no more than 4 hours each day, 15 hours / 4 hours each day = 3.75
They will need to plan to study 4 days to allow for all the time they need.
The answer is 4\\\\

Below is an instruction that describes a task, paired with an input that provides further context. Write a response that appropriately completes the request.\\\\\#\#\# Instruction:\\Given a problem scenario with numerical data, perform the necessary calculations and provide a detailed step-by-step solution, ending the response with 'The answer is'.\\\\\#\#\# Input:\\Mark's basketball team scores 25 2 pointers, 8 3 pointers and 10 free throws. Their opponents score double the 2 pointers but half the 3 pointers and free throws.  What's the total number of points scored by both teams added together?
\\\#\#\# Response:
Let's think step by step. Mark's team scores 25 2 pointers, meaning they scored 25*2= 50 points in 2 pointers.
His team also scores 6 3 pointers, meaning they scored 8*3= 24 points in 3 pointers
They scored 10 free throws, and free throws count as one point so they scored 10*1=10 points in free throws.
All together his team scored 50+24+10= 84 points
Mark's opponents scored double his team's number of 2 pointers, meaning they scored 50*2=100 points in 2 pointers.
His opponents scored half his team's number of 3 pointers, meaning they scored 24/2= 12 points in 3 pointers.
They also scored half Mark's team's points in free throws, meaning they scored 10/2=5 points in free throws.
All together Mark's opponents scored 100+12+5=117 points
The total score for the game is both team's scores added together, so it is 84+117=201 points
The answer is 201\\\\

Below is an instruction that describes a task, paired with an input that provides further context. Write a response that appropriately completes the request.\\\\\#\#\# Instruction:\\Given a problem scenario with numerical data, perform the necessary calculations and provide a detailed step-by-step solution, ending the response with 'The answer is'.\\\\\#\#\# Input:\\Bella has two times as many marbles as frisbees. She also has 20 more frisbees than deck cards. If she buys 2/5 times more of each item, what would be the total number of the items she will have if she currently has 60 marbles?
\\\#\#\# Response:
Let's think step by step. When Bella buys 2/5 times more marbles, she'll have increased the number of marbles by 2/5*60 = 24
The total number of marbles she'll have is 60+24 = 84
If Bella currently has 60 marbles, and she has two times as many marbles as frisbees, she has 60/2 = 30 frisbees.
If Bella buys 2/5 times more frisbees, she'll have 2/5*30 = 12 more frisbees.
The total number of frisbees she'll have will increase to 30+12 = 42
Bella also has 20 more frisbees than deck cards, meaning she has 30-20 = 10 deck cards
If she buys 2/5 times more deck cards, she'll have 2/5*10 = 4 more deck cards.
The total number of deck cards she'll have is 10+4 = 14
Together, Bella will have a total of 14+42+84 = 140 items
The answer is 140\\\\

Below is an instruction that describes a task, paired with an input that provides further context. Write a response that appropriately completes the request.\\\\\#\#\# Instruction:\\Given a problem scenario with numerical data, perform the necessary calculations and provide a detailed step-by-step solution, ending the response with 'The answer is'.\\\\\#\#\# Input:\\A group of 4 fruit baskets contains 9 apples, 15 oranges, and 14 bananas in the first three baskets and 2 less of each fruit in the fourth basket. How many fruits are there?
\\\#\#\# Response:
Let's think step by step. For the first three baskets, the number of apples and oranges in one basket is 9+15=24
In total, together with bananas, the number of fruits in one basket is 24+14=38 for the first three baskets.
Since there are three baskets each having 38 fruits, there are 3*38=114 fruits in the first three baskets.
The number of apples in the fourth basket is 9-2=7
There are also 15-2=13 oranges in the fourth basket
The combined number of oranges and apples in the fourth basket is 13+7=20
The fourth basket also contains 14-2=12 bananas.
In total, the fourth basket has 20+12=32 fruits.
The four baskets together have 32+114=146 fruits.
The answer is 146\\\\

Below is an instruction that describes a task, paired with an input that provides further context. Write a response that appropriately completes the request.\\\\\#\#\# Instruction:\\Given a problem scenario with numerical data, perform the necessary calculations and provide a detailed step-by-step solution, ending the response with 'The answer is'.\\\\\#\#\# Input:\\\{question\}\\\\\#\#\# Response:
Let's think step by step.\\
\toprule
\textbf{HumanEval} \\
\midrule
Below is an instruction that describes a task, paired with an input that provides further context. Write a response that appropriately completes the request.
\\\\
\#\#\# Instruction:\\
Complete the following python code.
\\\\
\#\#\# Input:\\
Check if in given list of numbers, are any two numbers closer to each other than given threshold.
\\\(>>>\) has\_close\_elements([1.0, 2.0, 3.0], 0.5)
False
\\\(>>>\) has\_close\_elements([1.0, 2.8, 3.0, 4.0, 5.0, 2.0], 0.3)
True
\\\\
\#\#\# Response:\\
from typing import List\\\\

def has\_close\_elements(numbers: List[float], threshold: float) -> bool:\\
\ \ \  """ Check if in given list of numbers, are any two numbers closer to each other than
    given threshold.
\\\ \ \     \(>>>\) has\_close\_elements([1.0, 2.0, 3.0], 0.5)
    False
 \\\ \ \    \(>>>\) has\_close\_elements([1.0, 2.8, 3.0, 4.0, 5.0, 2.0], 0.3)
    True
    """\\
\bottomrule
\end{xtabular*}
\captionof{table}{The templates used for evaluation.}
\label{tab:temp_eval}
\end{small}
\end{center}

\end{document}